\documentclass{bmvc2k}
\usepackage[normalem]{ulem}
\usepackage{graphicx}
\usepackage[export]{adjustbox}
\usepackage{duckuments}
\usepackage{tikz}
\usepackage{multirow}
\usepackage{blindtext}
\usepackage{floatrow}
\newfloatcommand{capbtabbox}{table}[][\FBwidth]
\usepackage{enumitem}
\usepackage{booktabs}
\usepackage{subfigure}
\newcommand*{\Comb}[2]{{}^{#1}C_{#2}}%

\def\checkmark{\tikz\fill[scale=0.4](0,.35) -- (.25,0) -- (1,.7) -- (.25,.15) -- cycle;} 

\newcommand{\vidimwidth}{0.2}
\newcommand{\vidfolder}{videos2}

\title{Livestock Monitoring with Transformer}

\newcommand\blfootnote[1]{%
  \begingroup
  \renewcommand\thefootnote{}\footnote{#1}%
  \addtocounter{footnote}{-1}%
  \endgroup
}

\addauthor{Bhavesh Tangirala$^\ast$}{bhaveshtangirala786@gmail.com}{1}
\addauthor{Ishan Bhandari$^\ast$}{bhandari.ishan19@gmail.com}{1}
\addauthor{Daniel Laszlo}{daniel@serket-tech.com}{2}
\addauthor{Deepak K. Gupta}{deepak.gupta@iitism.ac.in}{1}
\addauthor{Rajat M. Thomas}{rajatthomas@gmail.com}{2}
\addauthor{Devanshu Arya}{d.arya@uva.nl}{23}

\addinstitution{
 Transmute AI Lab\\
 Indian Institute of Technology\\
 ISM Dhanbad, India
}
\addinstitution{
 Serket-Tech\\
 The Netherlands
}
\addinstitution{
 University of Amsterdam\\
 The Netherlands
}

\runninghead{Starformer}{Livestock Monitoring with Transformer}

\begin{document}

\maketitle
% This creates the footnote text
\blfootnote{$^\ast$ Equal Contribution.}

\begin{abstract}

 Tracking the behaviour of livestock enables early detection and thus prevention of contagious diseases in modern animal farms. Apart from economic gains, this would reduce the amount of antibiotics used in livestock farming which otherwise enters the human diet exasperating the epidemic of antibiotic resistance {\textemdash} a leading cause of death. We could use standard video cameras, available in most modern farms, to monitor livestock. However, most computer vision algorithms perform poorly on this task, primarily because, (i) animals bred in farms look identical, lacking any obvious spatial signature, (ii) none of the existing trackers are robust for long duration, and (iii) real-world conditions such as changing illumination, frequent occlusion, varying camera angles, and sizes of the animals make it hard for models to generalize. Given these challenges, we develop an end-to-end behaviour monitoring system for group-housed pigs to perform simultaneous instance level segmentation, tracking, action recognition and re-identification (STAR) tasks. We present \textsc{starformer}, the first end-to-end multiple-object livestock monitoring framework that learns instance-level embeddings for grouped pigs through the use of transformer architecture. For benchmarking, we present \textsc{Pigtrace}, a carefully curated dataset comprising video sequences with
instance level bounding box, segmentation, tracking and activity classification of pigs in real indoor farming environment. Using simultaneous optimization on STAR tasks we show that \textsc{starformer} outperforms popular baseline models trained for individual tasks.

\end{abstract}

%-------------------------------------------------------------------------
%%%%%%%%% BODY TEXT
\section{Introduction}
In livestock farming, contagious diseases among animals cause havoc to their well-being and massive economic damage to the farmer. As a precaution, farms excessively use veterinary antibiotics leading to soil pollution and increased antibiotic resistance in humans ~\cite{wallinga2018better}. To prevent this indiscriminate use of antibiotics, diseases need to be identified at an early stage. Tracking animal-behaviour (movements and feeding patterns) enables us to do so~\cite{taylor1986pig, wedin2018early}. Obvious visible signs of sickness almost always occur when the disease is in an advanced stage and has already spread to the rest of the herd. Adding to the problem, a typical farmer has about a second to visually inspect an individual \cite{o2012assessment}. Thus, continuous monitoring of animal behaviour to detect anomalies as early as possible is invaluable to livestock \mbox{farming \cite{frost1997review}}. Although, the techniques discussed in this paper are applicable across different livestock, we present all our investigation based on pig livestock farming which is one of the most widespread form of livestock in world \footnote{http://www.fao.org/faostat/en/}, and prone to deadly infections.

For mid- or large-size farms continuous tracking of individual animals is not humanly possible. For instance, only two seconds of daily observation per pig is recommended in modern swine facilities \citep{america2014standard}. 
% Such an analysis is laborious, non-scalable and time sensitive. 
Although the use of radio identification devices provide tracking data on every individual animal, there are several disadvantages to methods that rely on the use of wearable equipment such as being invasive, prone to damage, expensive and are often the cause of infections in farm animals ~\mbox{\cite{schleppe2010challenges,neethirajan2017recent}}. Stationary RGB cameras, already available in most modern farms, facilitate implementation that is cost effective, non-invasive, and scalable. Such installations can work virtually in any indoor farming environment and does not require extensive maintenance. With the advent of precision livestock farming using computer vision techniques, continuous monitoring of livestock by video surveillance has been a growing field of study ~\cite{kashiha2013automatic, berckmans2014precision, psota2019multi}.   

Several works have been proposed in the recent years focusing on  action classification, detection and segmentation problems in livestock \cite{oczak2014classification,nilsson2015development, nasirahmadi2017implementation, ter2018beef}. Most of these works rarely consider more than one task in a single model. To our knowledge there exists no work that provides a single robust behaviour analysis model which can perform the STAR tasks i.e. segmentation, tracking, action recognition and re-identification. We argue that for a complete computer vision based livestock behaviour monitoring system, an end-to-end framework is needed which can take into account these STAR tasks simultaneously. The advantage of such a unified model would be that the learning process for multiple tasks positively affects each other in building robust and generalizable low-level representations of the image/video, thereby reducing the error to an extent beyond what can be achieved through individual training on each task. Based on this hypothesis, we present \textsc{starformer}, an end-to-end domain-adaptive transformer based model for segmentation, tracking, action recognition and re-identification (STAR) of livestock in closed environments.

% Continuous pig detection, tracking and their behaviour analysis  is a  significant task for improving the current animal farming. It can lead to a notable decrease in medicine, antibiotic usage, as vets/farmers can start medical treatments at an early stage before serious symptoms occur. Further, reducing the use of antibiotics and their negative impacts on the environment. It can also help in decreasing the number of aggressive events among the pigs, as the analysis of aggressive events helps to identify those pigs which are usually involved in these events. Several works have been proposed (cite[Nasirahmadi, Edwards,Sturm, 2017Ott et al., 2014Viazzi et al., 2014]) in the recent years focussing on pig detection and their segmentation. However, almost all these methods do not provide a single robust behaviour analysis model which can perform pig tracking, activity classification and anomaly detection.

% {\color{red}Mention how the problem of livestocks is different from other multiple-object problems - closed environment, very similar objects.}
While common tasks like pedestrian detection and object tracking lend themselves well to pre-trained networks and existing datasets, there exist unique challenges when monitoring livestock in a video. Pig monitoring in closed farming environment, in particular, poses the hard computer vision challenges of confusion between different pigs due to visual similarity, abrupt motions due to aggressive behaviour of pigs, frequent occlusions, huddling of pigs on top of each other, among others. These issues are seldom seen in traditional video datasets. Moreover, livestock activities are  confined, extremely repetitive and cyclic over time which makes them different from traditional action datasets. To build robust models that can tackle the issues outlined above, there is a need to acquire custom datasets and accompanying solutions. Any such dataset should capture the variability in conditions and livestock's behaviour must be annotated under the supervision of expert animal scientists. 

Transformers \cite{carion2020end, zhu2020deformable} have shown great success in a wide range of domains including natural language \cite{devlin2018bert}, images \cite{carion2020end}, video \cite{sun2020transtrack} and audio \cite{wang2020transformer}. In the visual domain, transformers have achieved promising results on object detection, panoptic segmentation and multi-object tracking (MOT) \cite{sun2020transtrack}. Recently, several Transformer-based tracking approach \cite{meinhardt2021trackformer,zhao2021trtr} has tried to exploit the advantages of the encode-decoder architecture which can encode frame-level features from a convolutional neural network (CNN)  \cite{he2016deep}  and decodes queries into bounding boxes associated with identities. We leverage this paradigm  by devising an end-to-end framework using the learned embeddings for performing STAR tasks.

In this paper, we present \textsc{starformer}, a domain-adaptive transformer-based model for simultaneous segmentation, tracking, action recognition and re-identification (STAR) of livestock for robust behavioural monitoring in closed environments. Starformer employs a \emph{spatio-temporal contrastive loss} term that stresses on learning the intra-object temporal similarity as well as the inter-object differences. We demonstrate that our formulation of a multi-objective optimization problem, promoting \emph{multi-task assistance during training},  enriches the feature representations significantly allowing to make clear distinctions between visually identical objects. For benchmarking, we further present \textsc{PigTrace}, a dataset of 30 videos containing multiple pigs to benchmark performance of algorithms for multiple-object tracking and action recognition in closed environment. Numerical experiments on \textsc{Pigtrace} and another large-scale dataset show \textsc{starformer} outperforming the competitive tracking baselines.

\begin{figure}[t]
\begin{tabular}{lllll}

\includegraphics[width=\vidimwidth\linewidth,valign=m]{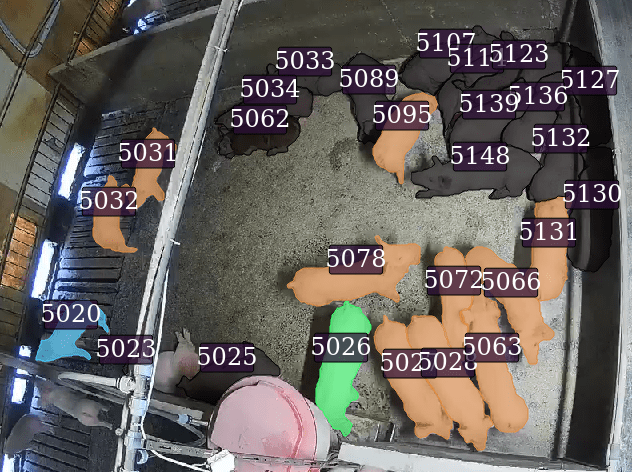} &
\includegraphics[width=\vidimwidth\linewidth,valign=m]{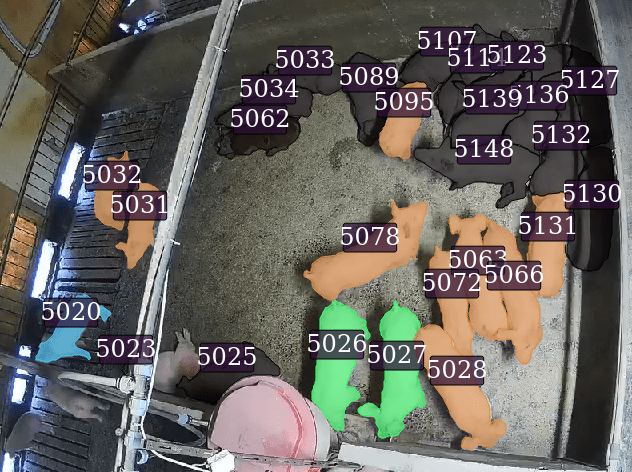} &
\includegraphics[width=\vidimwidth\linewidth,valign=m]{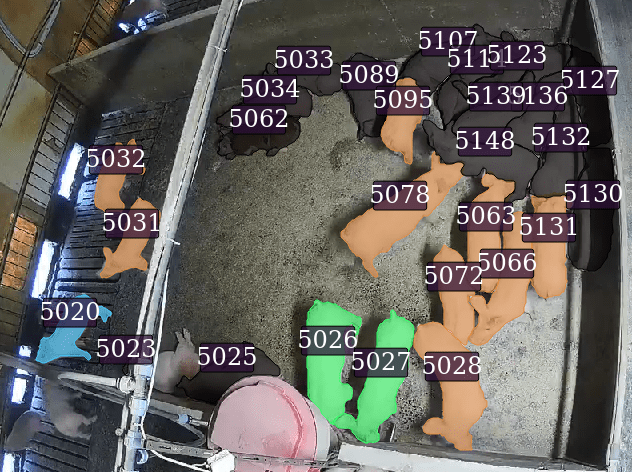} &
\includegraphics[width=\vidimwidth\linewidth,valign=m]{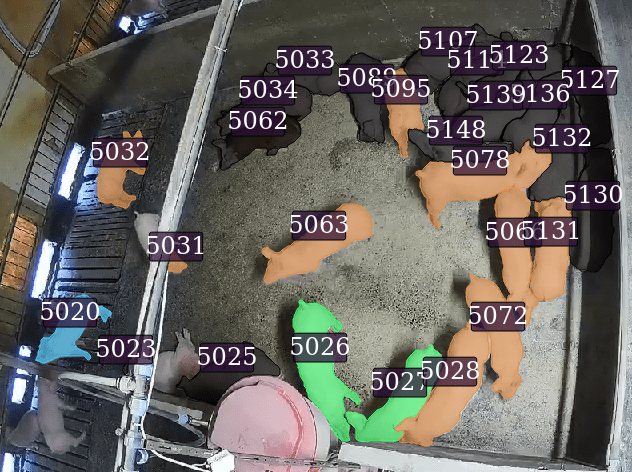}\vspace{0.5em}\\
\includegraphics[width=\vidimwidth\linewidth,valign=m]{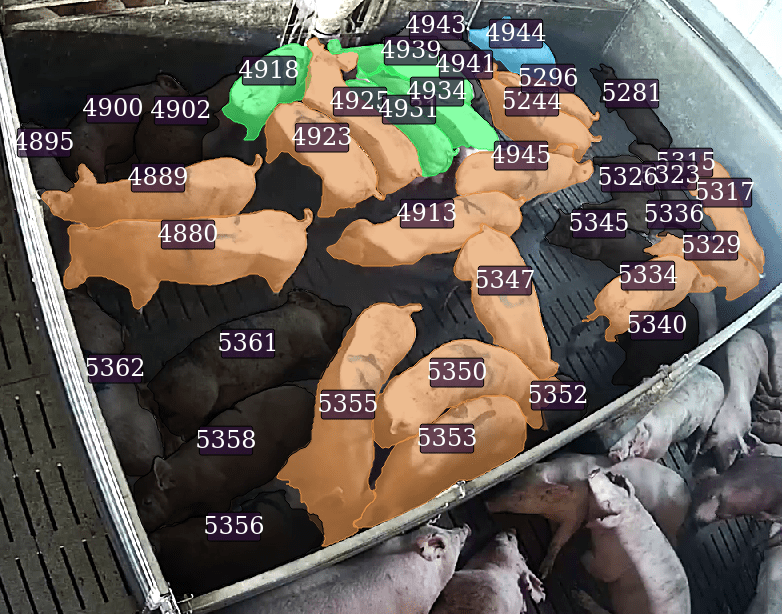} &
\includegraphics[width=\vidimwidth\linewidth,valign=m]{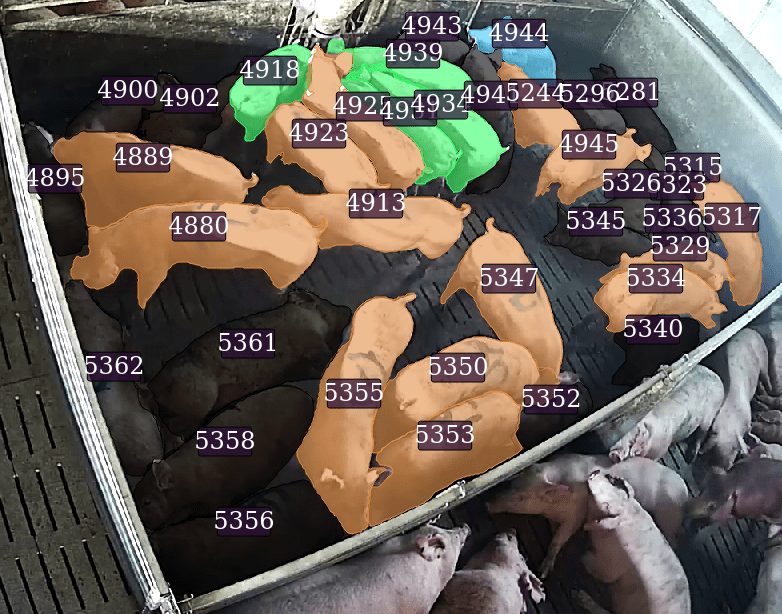} &
\includegraphics[width=\vidimwidth\linewidth,valign=m]{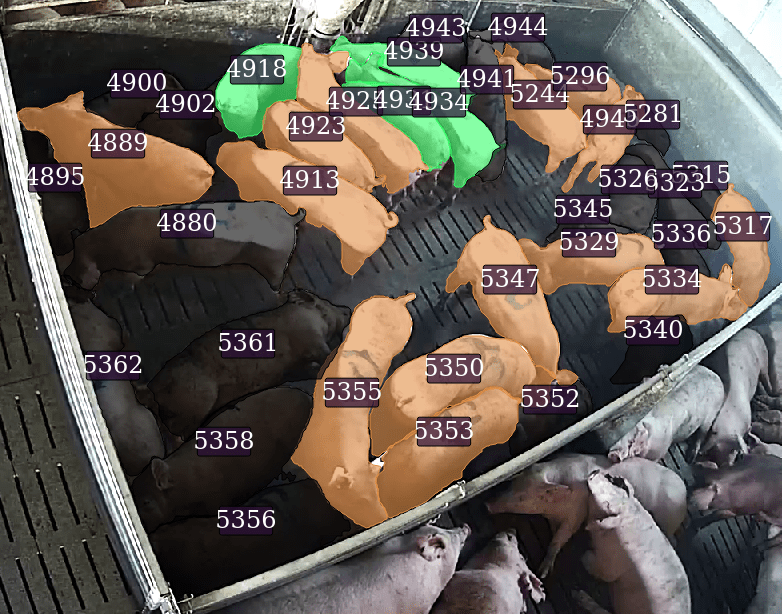} &
\includegraphics[width=\vidimwidth\linewidth,valign=m]{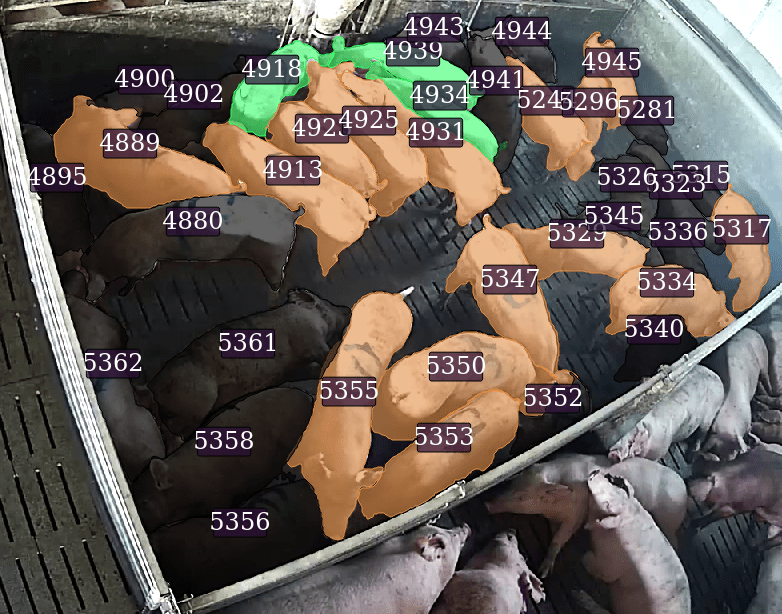}\vspace{0.5em}\\
\includegraphics[width=\vidimwidth\linewidth,valign=m]{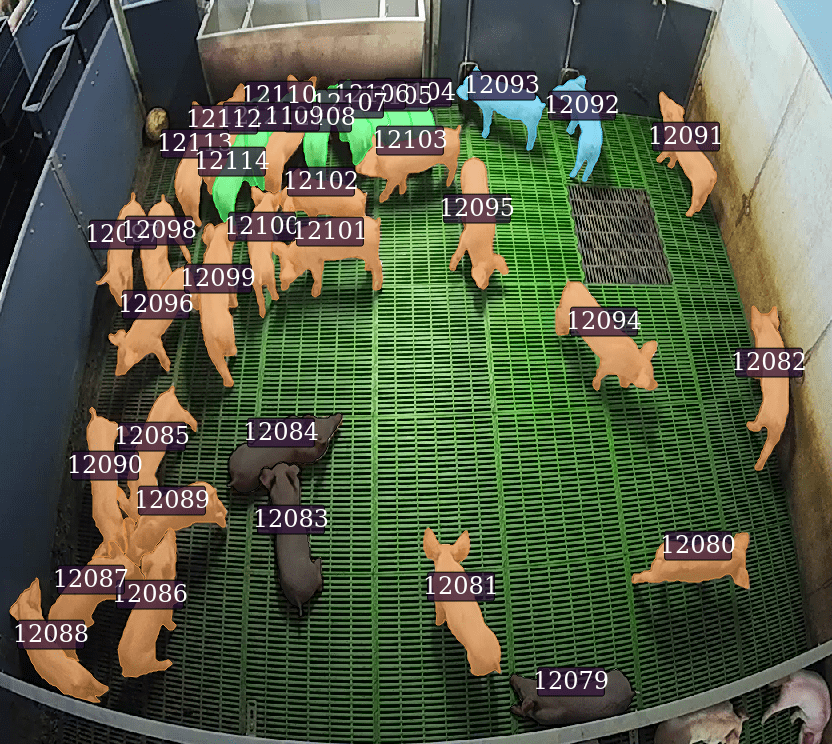} &
\includegraphics[width=\vidimwidth\linewidth,valign=m]{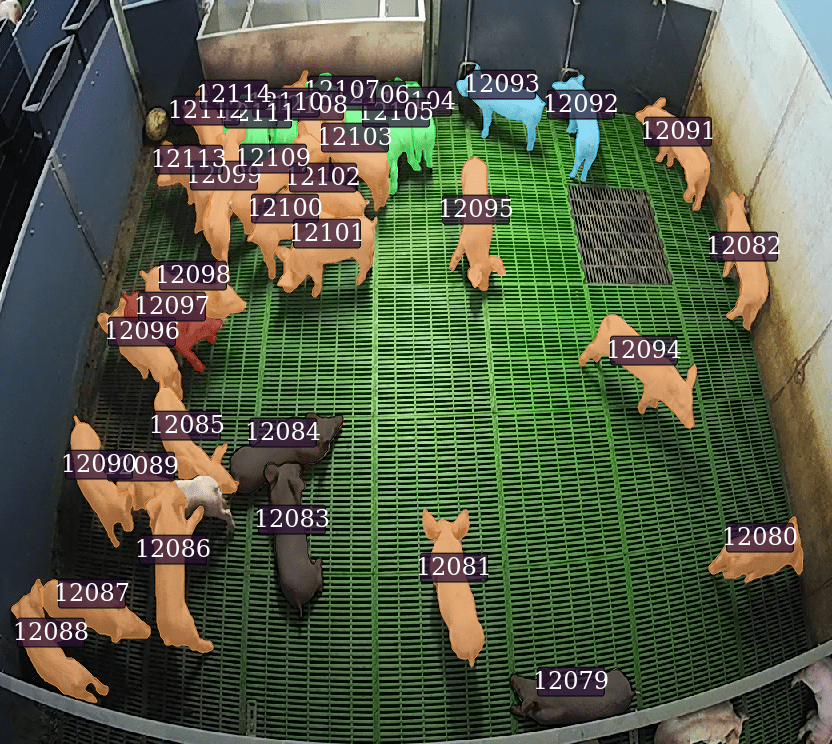} &
\includegraphics[width=\vidimwidth\linewidth,valign=m]{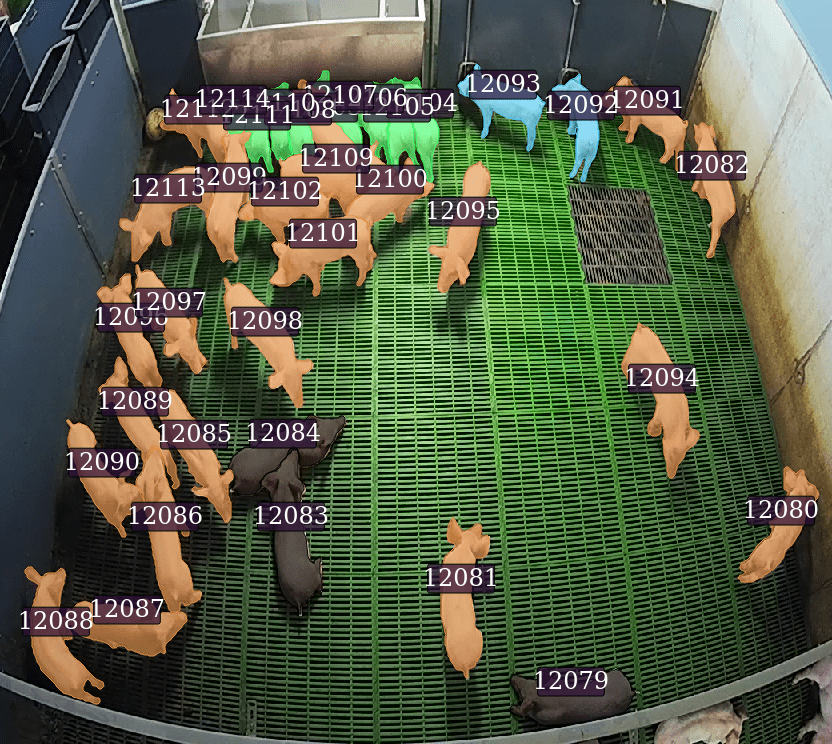} &
\includegraphics[width=\vidimwidth\linewidth,valign=m]{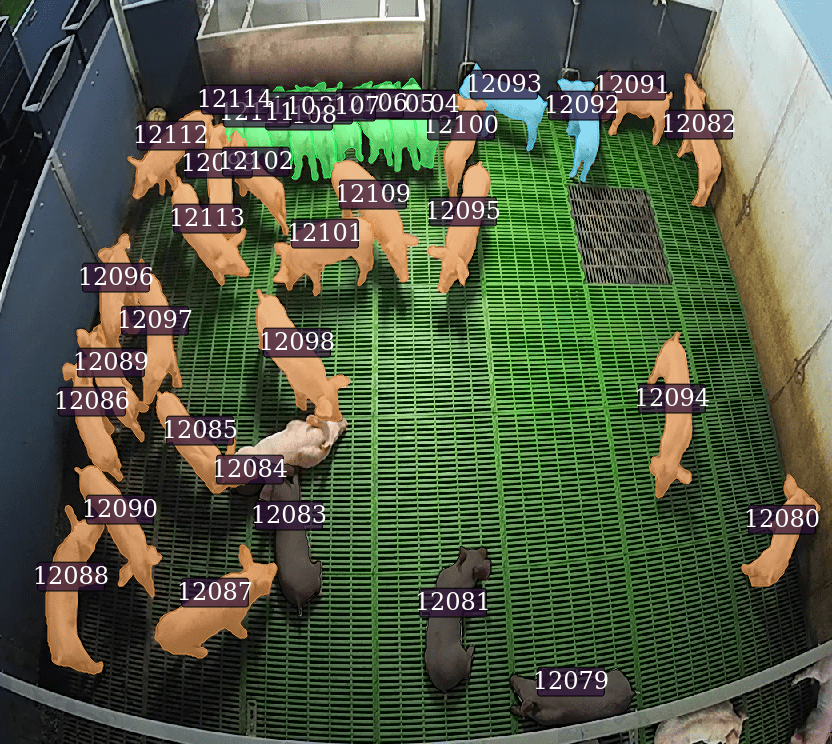}\\
\end{tabular}
\caption{Example frames and annotations from 3 videos of the \textsc{Pigtrace} dataset. Each row corresponds to a video from a different farm, and each column to successive annotated frames. The colors denote animal actions: green -- eating; blue -- drinking; red -- aggression; orange -- walking/standing; black -- inactive. Numbers on the animals are their unique ID.}
\end{figure}

\section{Related Work}
behaviour monitoring for livestocks has been a topic of research over the past two decades \cite{magee2000}, researchers have approached this problem from multitutde of different angles \cite{kim2010animal}. These include 3D tracking via wearable ultra-wide band (UWB) devices \cite{giancola2005uwb,porto2014localisation}, GPS \cite{schwager2007robust, kim2010animal}, inertial measurement unit (IMU) activity trackers \cite{mayer2004cattle,ruiz2009review,escalante2013sow,alvarenga2016using} and RFID ear tags \cite{voulodimos2010complete, feng2013development, floyd2015rfid}. Although livestock monitoring methods using radio identification devices directly provide data on individual animals, using wearables have several practical disadvantages \cite{neethirajan2017recent}. In contrast, video-based approaches \cite{mittek2016health,chen2020siambomb} provides cheaper and information-rich data to identify precisely what each animal is doing at all times. However, lack of robust approach for discerning identical looking animals and expensive data acquisition makes it hard to devise a unified vision based livestock monitoring framework. 

Visual detection of multiple moving targets with a static camera often begins with segmentation of foreground objects followed by background subtraction. Traditional computer vision methods such as Mask R-CNN \cite{he2016deep} uses top-down approach for performing instance-level object segmentation and keypoint detection. However, it can not efficiently identify unique instances if sufficient separation between the target does not exists such as in the case of group-housed animals. This is because it relies on \emph{a priori} region proposal, making it inherently unable to separate objects with significant bounding box overlap \cite{psota2019multi}.  Recently, several approaches have been proposed for tracking animals in closed environments which includes bottom-up keypoint detection for cow tracking \cite{ardo2018convolutional}, using three dimensional video cameras (top view with depth sensors) \cite{kulikov2014application,stavrakakis2015validity,kim2010animal}, using  Gaussian mixture models for pig tracking  \cite{ahrendt2011development} or using a computationally heavy implementation of Faster RNN with bounding box regression and segmentation masks \cite{wu2019multi}. 
Methods exist to identify the lying
behaviour of group-housed pigs as a function of temperature
\cite{nasirahmadi2016automatic} and the movement patterns of individual pigs and the
entire herd have been extracted through optical flow to detect
abnormal behaviours \cite{gronskyte2016monitoring}. Most of these approaches are either computationally expensive during inference, lack robustness to change in environment condition and occlusion or are incapable of identifying individual animals. Thus, current approaches fail to mitigate the problems of farmers which includes continuous real-time behaviour mentoring and re-identification of animals.

\section{\textsc{Pigtrace} Dataset}
To encourage researchers to participate and benchmark their approaches against this challenging task, we provide a unique dataset we call \textsc{Pigtrace} of videos from real-world animal farms along with detailed instance-level mask and action annotations.

\textsc{Pigtrace} consists of around 30 video sequences collected from five different farms in Europe. Each video is about five seconds long, and we are able to identify typical behaviours such as eating, drinking, laying, standing and walking in these videos. Seven videos are annotated at 6 frames-per-second (FPS) resulting in 30 annotated frames per video, and 23 other videos are annotated at 3 FPS resulting in 15 frames per video. In total there are 540 frames in this dataset. The dataset also includes all the annotation which are instance (pig) mask for each animal, a unique ID associated with each animal in a video along with a label of the actions the animals are performing. Code will also be provided to ease the use of this dataset on custom models along with APIs for evaluation metrics.

\section{Proposed Method}
\subsection{\textsc{Starformer}}
% {\color{red} \textbf{Inspired by detr and high level description of using embeddings for different heads}}
\textsc{Starformer} is a multi-task transformer model designed to perform segmentation, tracking, action recognition and re-identification (STAR) in livestock by extending the popular DETR  object detection model \cite{carion2020end, zhu2020deformable}. DETR introduced the concept of \emph{object queries} -- a fixed number of learned positional embeddings. These embeddings can be extracted as representations for possible object instances in an image. Motivated by this, \textsc{starformer} extends DETR to learn individual embeddings that are more discerning of an instance via the STAR multi-task learning.
%proposes a multi-task learning framework with 4 training heads, namely, detection, segmentation, action and tracking to perform end-to-end STAR tasks.

The base model for \textsc{starformer} is a DETR model with detection and segmentation heads pretrained on COCO detection and panoptic segmentation dataset. Since pigs are not part of the 80 classes of the COCO dataset, we trained a new classification head through re-training the base DETR \footnote{https://github.com/facebookresearch/detr} architecture by unfreezing both its encoder and decoder. 

\begin{figure}[htbp]
    \centering
    \includegraphics[scale=0.43]{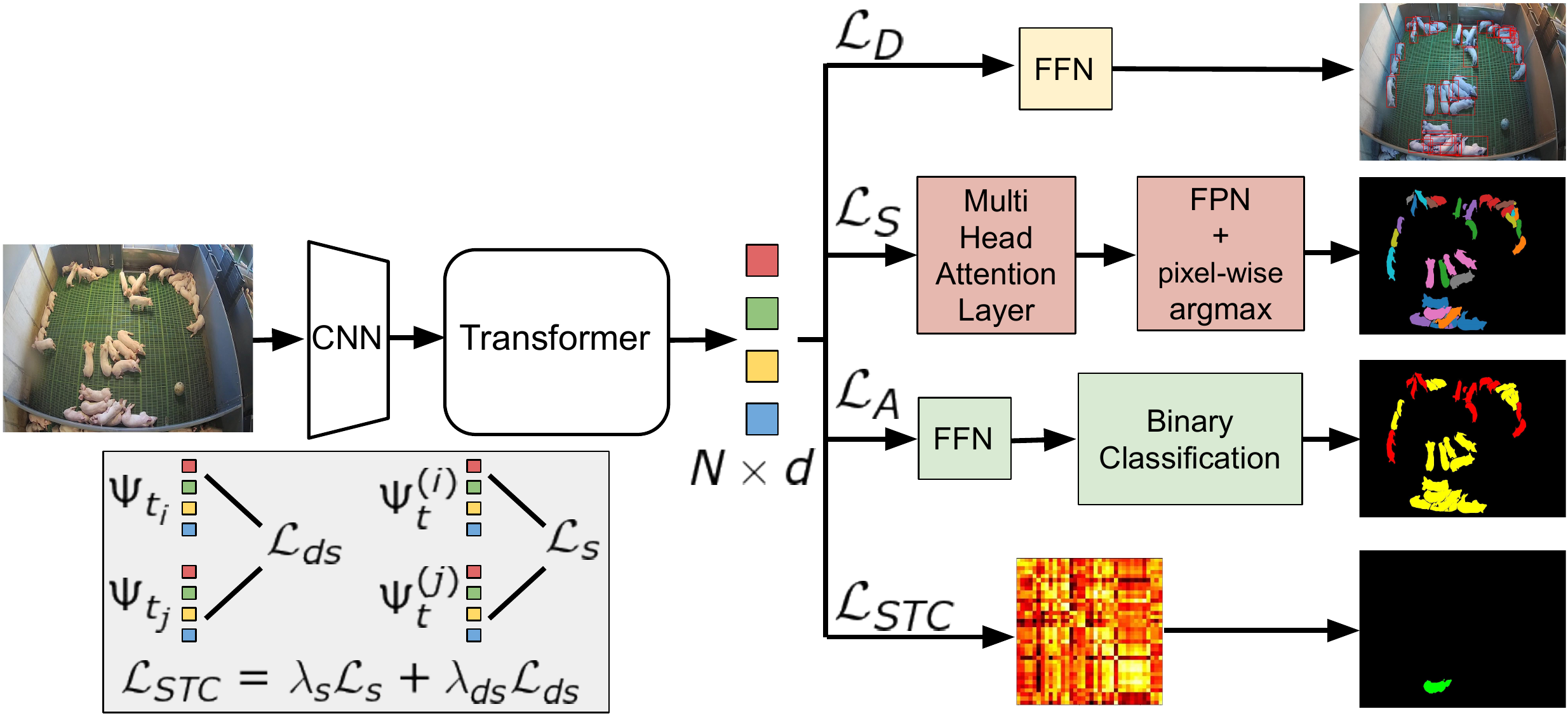}
    \caption{Schematic representation of \textsc{starformer} typically designed for livestock monitoring. The four losses corresponding to the four heads namely detection ($\mathcal{L_D}$), segmentation ($\mathcal{L_S}$), action ($\mathcal{L_A}$) (red - active, yellow - inactive) and spatio-temporal contrastive losses $\mathcal ({L_{STC}}$) are shown.}
    \label{main_fig}
\end{figure}

Figure \ref{main_fig} represents the architecture of \textsc{starformer} -- a ResNet-101 backbone followed by a transformer comprising 6 encoder-decoder layers with fixed size positional encodings (object queries). 
% and an action classification head (built on top of the decoder).
Based on \emph{a priori} knowledge of the number of pigs, the transformer module generates $N$ latent embeddings, each corresponding to an individual pig. The key idea of  \textsc{starformer} is to improve the embeddings by designing four heads, each optimizing a loss function of the STAR tasks.

For segmentation, \textsc{starformer} uses a multi-head attention layer and a feature pyramid network (FPN) - style CNN.
% \dpk{Expand on intuitive understanding of such strategy for segmentation - why it helps. If DETR used it, what was the reason that they used it.}
The detection head consists of a Feed Forward Network (FFN) that is a 3-layer perceptron with ReLU activation function and a linear projection layer. The detection head FFN predicts a bounding box, and the linear layer assigns a label to each pig. Actions are detected by parsing the instance level embeddings through another FFN which in turn augments the instance level embeddings (output of Decoder), to classify each object (pigs) into "Active" (standing) or "Inactive" (sitting/lying) classes. As shown in figure \ref{main_fig}, for the tracking head, we devise a spatio-temporal contrastive training approach which aims to increase the  similarity of an individual pig across the temporal direction while making sure the embeddings for pigs within the same frame are dissimilar. To further enhance these embeddings for long-term pig re-identification, we extend 
the spatio-temporal contrastive training approach on non-continous frames.  Frames are taken pairwise from a batch of $K$ frames, resulting in $\Comb{K}{2}$ possible combinations. Such a training strategy can extract motion patterns and shape variations in pigs, making the model to implicitly learn individual representations even for long-term scenarios.

% The key feature of Transformer is it’s self-attention mechanism which allows us to extract global context of the entire image as well as focus on the specific objects that are predicted. It explicitly models all interactions of two elements taken at once from a sequence, which helps in overcoming the problem of duplicate predictions. It was observed that the encoder helps separate the instances to form a better set of features and makes the localisation task easier for the decoder which refines the object representations.

% \subsection{Task Augmented Instance Level Embeddings}

\subsection{Multi-objective formulation for embedding enrichment}

We discuss here briefly the loss functions associated with the different heads of our \textsc{starformer} network.

% \subsection{Training losses}

\textbf{Detection loss --} following the DETR strategy, we employ the Hungarian loss $\mathcal{L}_D$ \cite{carion2020end}, but with only one class (pigs). This loss primarily combines the classification loss (cross-entropy loss training the model to classify as pig or background) and the bounding box loss (linear combination of L1 loss, and generalised IoU loss).

\textbf{Segmentation loss --}  we pass the feature embeddings to the instance segmentation head, and simply use an \emph{argmax} over the mask scores at each pixel, and assign the corresponding categories to the resulting masks.
The final resolution of the masks has stride of four and each mask is supervised independently using the DICE/F-1 loss \cite{milletari2016fully} and Focal loss \cite{lin2017focal}.

\textbf{Spatio-Temporal Contrastive Loss --} to ensure that our tracking model works well against the strong visual similarity among the pigs, we introduce a customized contrastive loss term that trains the model to better differentiate between the multiple pigs within the same frame, as well as improves the motion flow across subsequent frames for any individual pig. To compute the spatio-temporal contrastive loss $\mathcal{L}_{STC}$, we use the embeddings $\psi_t^{(i)}$ obtained from the last decoder layer for every individual pig from two closely spaced frames of the video. Here, $i \in N$ denotes the index of the pig, and $N$ denotes the total number of pigs as well as the number of embeddings per frame. We define $\mathcal{L}_{STC}$ as 
\begin{equation}
    \mathcal{L}_{STC} = \lambda_s\mathcal{L}_s + \lambda_{ds}\mathcal{L}_{ds},
\end{equation}
where, $\mathcal{L}_s$ and $\mathcal{L}_{ds}$ denote similarity and dissimilarity loss terms, and $\lambda_s$ and $\lambda_{ds}$ are the respective weighting terms.

To compute the measures of similarity and dissimilarity, we employ the cosine distance metric. Further, the similarity loss $\mathcal{L}_{s}$ is computed for each frame individually, and for the $t^{\text{th}}$ frame, it can be stated as
\begin{align}
    \mathcal{L}_{s} = \sum_{i, j}\frac{\psi_t^{(i)} \cdot \psi_t^{(j)}}{\|\psi_t^{(i)}\|\|\psi_t^{(j)}\|} \enskip \forall \enskip {i, j}\in \{1, 2, \hdots, N\} \text{ and }i \neq j.
\end{align}
To compute $\mathcal{L}_{ds}$, we choose $\tau$ subsequent frames of the video and compute the loss for each frame pair for all $N$ objects or animals.
Based on this, we define
\begin{equation}
    \mathcal{L}_{ds} = \sum_{i=1}^N \sum_{t_1, t_2} \left(1 - \frac{\psi_{t_1}^{(i)} \cdot \psi_{t_2}^{(i)}}{\|\psi_{t_1}^{(i)}\|\|\psi_{t_2}^{(i)}\|}\right) \enskip \forall \enskip t_1, t_2 \in \tau \text{ and } t_1 \neq t_2.
\end{equation}

% This comprises of 2 loss : Classification loss and Bounding Box loss.
% Classification Loss - Cross entropy loss for training the classification head to classify objects as pig or background.Bounding Box loss - It is a linear combination of L1 loss, and generalised IoU loss which is scale-invariant and differentiable.

\textbf{Action loss.} We conjecture that basic activity such as  sitting or standing can help in augmenting the learned embeddings $\psi_t^{(i)}$ $\forall i \in N $ with useful information about a pig's shape and size. This is important as one of the most discerning factor in pigs are their shapes and sizes.  We place an action classification head which classifies each pig into 2 classes i.e. active (standing) or inactive (sitting) using a binary 
cross entropy loss, and is denoted as $\mathcal{L}_A$.

% We take features from the last decoder layer $(bs \times num queries \times dmodel)$. We hypothesise that each of these features contains information about the identity of the pig that it represents. We use this identity info to track pigs through a video by comparing these features. 
% Consider feature(t) and feature(t+1) are outputs of the last decoder layer in frame(t) and frame(t+1). 
% We perform matrix multiplication between these two features to get the matrix mul.
% Id loss consist of two parts - Similarity loss(Ls) and dissimilarity loss(Ld).
% Our goal is to minimise cosine distance b/w representations of same pigs across the frame, and minimize cosine similarity b/w representations of different pigs.

\section{Experiments}
 \textsc{Starformer} uses a ResNet-101 DETR model pre-trained on COCO dataset as backbone. We train \textsc{starformer} on 280 videos, each consisting of 15 frames in the format stated in Section 3. During training, the backbone DETR transformer is re-trained by unfreezing both the encoder and decoder and learning new set of object queries. 
 All experiments are run on 8 V100 GPUs.  Details on optimization can be found in the supplementary material. We evaluate the performance of \textsc{starformer} on each of the 4 STAR tasks on a validation set consisting  84 videos, with 15 frames each. In total there are 4200 frames in the training set and 1260 frames in the validation set. We further provide benchmark scores on the \textsc{Pigtrace} dataset. Table \ref{data_stats} summarizes the  training, validation and \textsc{Pigtrace} datasets.
 
\begin{table}[h]
\scalebox{0.9}{
\begin{tabular}{lcccc}
\hline
Dataset    & No. of Videos & Average frames  & Frame Rate (FPS) & Average Number of Pigs \\ \hline \hline
\textsc{Pigtrace}   & 30            & 18.6 $\pm$ 1.2           & 6      & 28.8 $\pm$ 8.8   \\
Training   & 280           & 15                       & 3      & 31.3 $\pm$ 8.7   \\
Validation & 84            & 15                       & 3      & 32.1 $\pm$ 9.2   \\ \hline
\end{tabular}
}
\caption{Table summarizing the \textsc{Pigtrace} dataset (publicly available) and the training and validation dataset used.}
\label{data_stats}
\end{table}

\subsection{Baseline and Evaluation Metrics}
To understand and benchmark how different heads of \textsc{starformer} contribute towards its performance, we introduce multiple baselines and evaluated metrices and we discuss them below with respect to the 4 tasks.

\textit{Segmentation.} The performance of \textsc{starformer} on instance level segmentation is compared with state-of-the-art implementation of  MaskR-CNN \cite{he2017mask} as in \cite{wu2019detectron2} and DETR whose decoder and object queries are fine-tuned using our training dataset. Note that, while evaluating \textsc{starformer} and DETR, we fix the number of predictions to be equal to the number of pigs in that video. This can be beneficial in livestock setting as the number of animals in closed environment will remain fixed over the course of a video. This constrains the model to not allow over or under predicting the number of embeddings. We report the mean average precision (mAP) over different IoU thresholds, from 0.5 to 0.95 (written as `0.5:0.95') and also mAP at 0.5 threshold. 

\textit{Tracking.} The segmentation masks obtained from the segmentation models are used to perform multi-object tracking by matching these masks temporally. Pig tracking is constrained such that the number of pigs remains same throughout the video. We use this constraint and fix the number of predictions to the number of pigs $N$ in the video which is known \emph{a priori}. For each video, we consider only the top $N$ predictions out of all object queries. Note that, we can also get an initial estimate of the number of pigs. This can be done in different ways such as by using the mode of the number of masks estimated over a period of time (burn-in period, before we start the tracking), which we have seen is a robust estimator of the number of pigs (pig counting). For the first frame, we form a one-to-one mapping between the ground truth instances with the predicted instances by greedily matching the pairs with maximum mask IoU at each step. Using this mapping and the mapping between the predicted instances across time frames, we match the ground truth instance of each frame with its corresponding predicted instance.

Although there are many methods available for tracking using segmentation masks or embeddings \cite{kuhn1955hungarian, bewley2016simple}, but in livestocks monitoring since the animals (there number and instances) are fixed, the camera is not moving and no animal leaves or enters the scene. These restrictions enables us to perform tracking with a rather straightforward matching strategy. For a proper analysis of how the tracking module performs, we use 2 different matching algorithms and compare the performances for each. Brief descriptions of these follow below.
\vspace{-1.5em}
\begin{enumerate}[noitemsep]
    \item \emph{Matching by mask.} Similarity between pigs is computed as IoU of their segmentation masks. We match the pigs by greedily matching the pair of pigs that exhibit highest IoU among all the pairs.
    \item \emph{Matching by Embedding.} To compute the extent of similarity between embeddings corresponding to different pigs, cosine distance measure is used. For every pig being matched, the distance in the Euclidean space should be less than $R$.
\end{enumerate}

We propose \emph{constrained multi-object tracking and segmentation accuracy} (cMOTSA) as a metric to evaluate problems of tracking with constraints. 
Due to the constraint of fixed count of livestock throughout the video, there will be no false negatives (FN) since 1-1 mapping exists now between the ground-truths and the respective instances obtained from prediction. 
We hope that this evaluation metric can accurately assess the capability of \textsc{starformer} in learning unique representations for each pig instance. It is defined as the ratio of the number of true positives \text{TP} (matched instance pairs with a mask IoU greater than 0.5) to all the positive predictions (|TP| + |FP|). False positives (FP) are the instance pairs with a mask IoU less than or equal to 0.5. Further, we also evaluate the tracking performance using scMOTSA, a soft variant of cMOTSA defined as $\text{scMOTSA} = \widetilde{\text{TP}} / (|\text{TP}| + |\text{FP}|)$, where $\widetilde{\text{TP}}$ denotes soft true positives. See details in the supplementary material. 
%Since it aggregates the soft true positives and not count the masks,  
% We also introduce a soft version of our metric soft cMOTSA (scMOTSA) as in {https://arxiv.org/abs/1902.03604}. We define TP~ as 
% TP~ = sigma(Iou())....

% We define scMOTSA as 
% scMOTSA = TP~ / (|TP| + |FP|)
%It aggregates the soft true positives TP~ rather than counting the number of masks with IoU > 0.5.

The standard evaluation metrics of MOTS, as stated in \cite{voigtlaender2019mots}, cannot be used for our study since these metrics require that there exists no overlap between masks of any two objects in the ground-truth as well as in the predictions. In other words, every pixel is allowed to be assigned to a maximum of one object. In our dataset, this is not the case and there occur frequent cases of pigs overlapping. Clearly, this property adds the instances of labelled occlusions in our dataset. Occlusion among the hard challenges of tracking \cite{gupta2021icpr, kuipers2020eccvw}, and we hope that model training on such datasets could also introduce invariance to a certain extent.

\textit{Action Classification.} The efficacy of \textsc{starformer} for the action classification task is evaluated through comparison with a ResNet-101 inspired model (Ac-ResNet) \cite{he2016deep} trained specifically to classify each pig into two classes - inactive (sitting) or active (standing). Details related to this baseline are provided in the supplementary material. We use area under the curve in receiver operating characteristic (AUC-ROC) curve as the evaluation metric.

\textit{Pig Re-Identification.}  
 We use Cumulative Matching Characteristics (CMC) scores \cite{gray2007evaluating} to compare re-identification between \textsc{starformer} and DETR. CMC curves are the most popular evaluation metrics for re-identification methods. CMC-$k$, also referred as Rank-$k$ matching accuracy, represents
the probability that a correct match appears in the top $k$
ranked retrieved results. Ranking, in our case is done by calculating embedding distances between pigs of different frame. CMC top-$k$ accuracy is 1 if correct match appears among the top $k$ values, else 0.
We plot CMC top-$k$ accuracies for discrete inter-frame intervals, i.e, the time interval between the two frames for which re-identification is being done.

\subsection{Results}
\begin{table}[h]
\centering
\scalebox{0.78}{
\begin{tabular}{lcccccccccc}
\toprule
 \multicolumn{1}{c}{\multirow{2}{*}{\textbf{Method}}} & \multicolumn{4}{c}{\textbf{Loss}}                                                                                               & \multicolumn{2}{c}{\textbf{mAP IoU:}}                           & \multicolumn{2}{c}{\textbf{Match masks}}    
 & \multicolumn{2}{c}{\textbf{Match embeddings}} \\
 \multicolumn{1}{c}{}                        & \multicolumn{1}{l}{$\mathcal{L}_D$}          & \multicolumn{1}{l}{$\mathcal{L}_S$} & \multicolumn{1}{l}{$\mathcal{L}_A$} & \multicolumn{1}{l}{$\mathcal{L}_{STC}$} & \multicolumn{1}{l}{0.5:0.95} & \multicolumn{1}{l}{0.5}  & \multicolumn{1}{c}{cMOTSA}  & \multicolumn{1}{c}{scMOTSA} 
 & \multicolumn{1}{c}{cMOTSA} &  \multicolumn{1}{c}{scMOTSA}\\ \midrule
                                       Mask R-CNN                                    &   -                                 &   -                        &  -                          &       -                     &       0.598                       &  0.860                                                    & 0.617 & -  & - &      -                      \\
                                       DETR                                         &   \checkmark                                  &       \checkmark                     &     -                      &       -                     &      0.600                        &        0.866                                              & 0.621 & 0.534  & 0.604     & 0.522                     \\
                                       \midrule

                                      \textsc{starformer}       & \checkmark          & \checkmark &        -                   &     -                       &                      0.663        &                         \textbf{0.920}                             & 0.743 & 0.642 & 0.714 & 0.611                    \\
                                                                             \textsc{starformer}       & \textbf{\checkmark} & \checkmark & \checkmark &     -                       &                          0.666    &               \textbf{0.920}                                       & 0.792 & 0.691 & 0.785 & 0.676
                                       \\
                                                                                 \textsc{starformer}       & \textbf{\checkmark} & \checkmark & \checkmark & \checkmark                    &                    \textbf{0.668}       &             \textbf{0.920}                                        & \textbf{0.805} & \textbf{0.704} &   \textbf{0.793} & \textbf{0.686}\\ \bottomrule 
\end{tabular}}
\caption{Performance scores for \textsc{starformer} and other baseline models for the tasks of segmentation and tracking obtained on validation set of pig livestock.}
\label{seg_track_result}
\end{table}

\begin{table}[h]
\scalebox{0.9}{
\begin{tabular}{lcccccccc}
\toprule
\multicolumn{1}{c}{\multirow{2}{*}{Method}} & \multicolumn{1}{c}{\textbf{Seg.} }                       & \multicolumn{1}{c}{\textbf{Track(M)}}                           &
\multicolumn{1}{c}{\textbf{Track(E)}}   &
\multicolumn{1}{c}{\textbf{Action}} & \multicolumn{3}{c}{\textbf{Re-Identify (CMC)}} \\
\multicolumn{1}{c}{}                        & \multicolumn{1}{c}{mAP}  & \multicolumn{1}{c}{cMOTSA}  & \multicolumn{1}{c}{cMOTSA} & \multicolumn{1}{c}{AUC}           & \multicolumn{1}{c}{R1}  & \multicolumn{1}{c}{R5} & \multicolumn{1}{c}{R10}             \\ \midrule
Ac-ResNet                                   &    -                                                    &    -  &      -              &           0.768                        &  -  & - & -                                     \\

Mask R-CNN                                   &     0.627                                                     &     0.550  &      -              &           -                        &  -  & - & -                                     \\
DETR                                        & 0.639                                                                              & 0.600  & 0.569                     &      -        & 0.678 & 0.846 &  0.904                                                         \\
\textsc{starformer}                                  &  \textbf{0.690}                                                                            &  \textbf{0.778 }      & \textbf{0.756}                 &     \textbf{0.985}                                 & \textbf{0.771}    & \textbf{0.895} & \textbf{0.939}                               \\ 
\bottomrule
\end{tabular}}
\caption{Performance scores obtained for \textsc{starformer} and the baseline models on the 4 STAR tasks. Here mAP is computed for 0.5:0.95. Further, Track(M) and Track(E) correspond to cases of matching by masks and matching my embeddings, respectively, and Action implies action recognition.}
\label{pigtrace}
\end{table}

We discuss here briefly the results of our experiments and present the important insights. 
Table \ref{seg_track_result} presents the results for segmentation and tracking of pigs on a validation set obtained with \textsc{starformer} as well as our baseline models. We observe that \textsc{starformer} consistently outperforms the two baseline models for all the evaluation metrices of segmentation and tracking. While the improvements for segmentation are approximately 6\%, absolute improvements of up to 20\% are observed for the task of segmentation. We also retrained our network with a Swin-Transformer backbone \cite{liu2021swin} and achieved a result of 0.76 mAP on the \textsc{PigTrace} dataset for the segmentation task. This was indeed a significant improvement in segmentation performance.
Further, for action classification, \textsc{starformer} obtains an AUC score of 0.98 compared to 0.742 obtained for the Ac-ResNet baseline. These results clearly demonstrate that training the model simultaneously over multiple tasks provides accurate performance over individual tasks themselves.
    
To further understand the effect of having multiple task heads, we also analyze a few cases where one of more task heads are removed from the original \textsc{starformer} model. These cases are also reported in Table \ref{seg_track_result}. As can be seen, the head with spatio-temporal contrastive loss when removed, has no adverse impact on segmentation performance but reduces the tracking performance by approximately 1\%. No change on segmentation is expected since contrastive loss primarily focuses on temporal flow of information in our case, while segmentation treats objects in every frame independent of each other. Similarly, when removing the action classification loss, tracking performance is significantly affected.

\begin{figure}[t]
\centering     %%% not \center
\subfigure[DETR Embeddings]{\label{fig:a}\includegraphics[width=60mm]{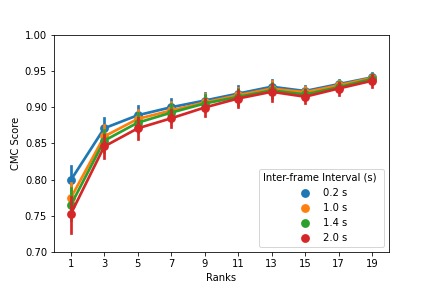}}
\subfigure[\textsc{Starformer} Embeddings]{\label{fig:b}\includegraphics[width=60mm]{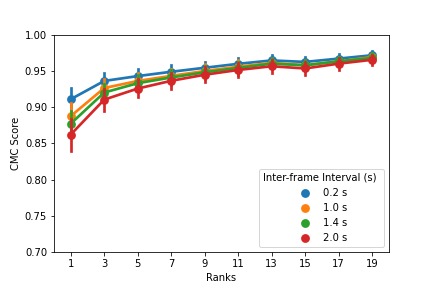}}
\vspace{-1em}
\caption{CMC curves for pig re-identifcation. Here, inter-frame interval implies the number of frames to be skipped to test the efficacy of re-identification, and rank $k$ implies the number of top predictions among which the desired target falls to be deemed as correct.}
\label{fig_reidentify}
\end{figure}

We further studied how well \textsc{starformer} performs for the task of re-identification and the results are presented in Fig. \ref{fig_reidentify}. We see that both DETR as well as \textsc{starformer} perform equally well for large values of $k$. However, large values of $k$ are not very suited for practical purposes, and performance at lower values of $k$ is more important. For lower values, we see that performance of DETR drops significantly for all choices of inter-frame intervals. On the contrary, \textsc{starformer} is more stable with very small drops for lower values of $k$. This implies that for long-term tracking,  \textsc{starformer} is expected to be more reliable.

\textsc{PigTrace.} We further analyzed the performance of \textsc{starformer} on \textsc{Pigtrace} dataset and the results are presented in Table \ref{pigtrace}. \textsc{Starformer} provides significant performance gains for all evaluation metrics across all the four STAR tasks.

\section{Conclusions and Future Scope}
In this paper, we presented \textsc{starformer}, a tranformer-based framework for behavioural monitoring of livestock. Using multi-task optimization,  \textsc{starformer}  outperforms  baseline methods for the individual tasks by significant margins. We further presented \textsc{Pigtrace}, the first benchmark dataset for behavioural monitoring of livestock in closed environment. We are working towards a semi-automated way of labelling to increase the volume of frames in the dataset. Initial results using the Swin-Transformer were promising and we continue to explore using the Swin-Transformer as the backbone for future research. Our current approach to tracking is rather simple. One clear research direction would be to incorporate modern data association methods between frames into our framework. For example, the constraints of livestock farming lends itself to the use of graph based tracking methods \cite{braso2020learning}. We hope that our proposed method along with the densely annotated dataset will pave the groundwork for future research and evaluation of methods for livestock monitoring.

\bibliography{egbib}

\begin{thebibliography}{56}
\providecommand{\natexlab}[1]{#1}
\providecommand{\url}[1]{\texttt{#1}}
\expandafter\ifx\csname urlstyle\endcsname\relax
  \providecommand{\doi}[1]{doi: #1}\else
  \providecommand{\doi}{doi: \begingroup \urlstyle{rm}\Url}\fi

\bibitem[Ahrendt et~al.(2011)Ahrendt, Gregersen, and
  Karstoft]{ahrendt2011development}
Peter Ahrendt, Torben Gregersen, and Henrik Karstoft.
\newblock Development of a real-time computer vision system for tracking
  loose-housed pigs.
\newblock \emph{Computers and Electronics in Agriculture}, 76\penalty0
  (2):\penalty0 169--174, 2011.

\bibitem[Alvarenga et~al.(2016)Alvarenga, Borges, Palkovi{\v{c}}, Rodina, Oddy,
  and Dobos]{alvarenga2016using}
FAP Alvarenga, I~Borges, L~Palkovi{\v{c}}, J~Rodina, VH~Oddy, and RC~Dobos.
\newblock Using a three-axis accelerometer to identify and classify sheep
  behaviour at pasture.
\newblock \emph{Applied Animal Behaviour Science}, 181:\penalty0 91--99, 2016.

\bibitem[America(2014)]{america2014standard}
PIC~North America.
\newblock Standard animal care: Daily routines.
\newblock \emph{Wean to Finish Manual; PIC: Hendersonville, TN, USA}, pages
  23--24, 2014.

\bibitem[Ard{\"o} et~al.(2018)Ard{\"o}, Guzhva, Nilsson, and
  Herlin]{ardo2018convolutional}
Hakan Ard{\"o}, Oleksiy Guzhva, Mikael Nilsson, and Anders~H Herlin.
\newblock Convolutional neural network-based cow interaction watchdog.
\newblock \emph{IET Computer Vision}, 12\penalty0 (2):\penalty0 171--177, 2018.

\bibitem[Berckmans(2014)]{berckmans2014precision}
Daniel Berckmans.
\newblock Precision livestock farming technologies for welfare management in
  intensive livestock systems.
\newblock \emph{Rev. Sci. Tech}, 33\penalty0 (1):\penalty0 189--196, 2014.

\bibitem[Bewley et~al.(2016)Bewley, Ge, Ott, Ramos, and
  Upcroft]{bewley2016simple}
Alex Bewley, Zongyuan Ge, Lionel Ott, Fabio Ramos, and Ben Upcroft.
\newblock Simple online and realtime tracking.
\newblock In \emph{2016 IEEE International Conference on Image Processing
  (ICIP)}, pages 3464--3468. IEEE, 2016.

\bibitem[Bras{\'o} and Leal-Taix{\'e}(2020)]{braso2020learning}
Guillem Bras{\'o} and Laura Leal-Taix{\'e}.
\newblock Learning a neural solver for multiple object tracking.
\newblock In \emph{Proceedings of the IEEE/CVF Conference on Computer Vision
  and Pattern Recognition}, pages 6247--6257, 2020.

\bibitem[Carion et~al.(2020)Carion, Massa, Synnaeve, Usunier, Kirillov, and
  Zagoruyko]{carion2020end}
Nicolas Carion, Francisco Massa, Gabriel Synnaeve, Nicolas Usunier, Alexander
  Kirillov, and Sergey Zagoruyko.
\newblock End-to-end object detection with transformers.
\newblock In \emph{European Conference on Computer Vision}, pages 213--229.
  Springer, 2020.

\bibitem[Chen et~al.(2020)Chen, Zhai, Liu, Li, Ding, Xie, and
  Han]{chen2020siambomb}
Xi~Chen, Hao Zhai, Danqian Liu, Weifu Li, Chaoyue Ding, Qiwei Xie, and Hua Han.
\newblock Siambomb: A real-time {AI}-based system for home-cage animal
  tracking, segmentation and behavioral analysis.
\newblock In \emph{IJCAI}, pages 5300--5302, 2020.

\bibitem[Devlin et~al.(2019)Devlin, Chang, Lee, and Toutanova]{devlin2018bert}
Jacob Devlin, Ming-Wei Chang, Kenton Lee, and Kristina Toutanova.
\newblock Bert: Pre-training of deep bidirectional transformers for language
  understanding.
\newblock \emph{Proceedings of NAACL-HLT}, 2019.

\bibitem[Escalante et~al.(2013)Escalante, Rodriguez, Cordero, Kristensen, and
  Cornou]{escalante2013sow}
Hugo~Jair Escalante, Sara~V Rodriguez, Jorge Cordero, Anders~Ringgaard
  Kristensen, and C{\'e}cile Cornou.
\newblock Sow-activity classification from acceleration patterns: a machine
  learning approach.
\newblock \emph{Computers and electronics in agriculture}, 93:\penalty0 17--26,
  2013.

\bibitem[Feng et~al.(2013)Feng, Fu, Wang, Xu, and Zhang]{feng2013development}
Jianying Feng, Zetian Fu, Zaiqiong Wang, Mark Xu, and Xiaoshuan Zhang.
\newblock Development and evaluation on a {RFID} based traceability system for
  cattle/beef quality safety in china.
\newblock \emph{Food control}, 31\penalty0 (2):\penalty0 314--325, 2013.

\bibitem[Floyd(2015)]{floyd2015rfid}
Raymond~E Floyd.
\newblock {RFID} in animal-tracking applications.
\newblock \emph{IEEE Potentials}, 34\penalty0 (5):\penalty0 32--33, 2015.

\bibitem[Frost et~al.(1997)Frost, Schofield, Beaulah, Mottram, Lines, and
  Wathes]{frost1997review}
AR~Frost, CP~Schofield, SA~Beaulah, TT~Mottram, JA~Lines, and CM~Wathes.
\newblock A review of livestock monitoring and the need for integrated systems.
\newblock \emph{Computers and electronics in agriculture}, 17\penalty0
  (2):\penalty0 139--159, 1997.

\bibitem[Giancola et~al.(2005)Giancola, Blazevic, Bucaille, De~Nardis,
  Di~Benedetto, Durand, Froc, Cuezva, Pierrot, Pirinen,
  et~al.]{giancola2005uwb}
Guerino Giancola, Ljubica Blazevic, Isabelle Bucaille, Luca De~Nardis, M-G
  Di~Benedetto, Yves Durand, Gwillerm Froc, Bego{\~n}a~Molinete Cuezva, J-B
  Pierrot, Pekka Pirinen, et~al.
\newblock {UWB MAC} and network solutions for low data rate with location and
  tracking applications.
\newblock In \emph{2005 IEEE International Conference on Ultra-Wideband}, pages
  758--763. IEEE, 2005.

\bibitem[Gray et~al.(2007)Gray, Brennan, and Tao]{gray2007evaluating}
Douglas Gray, Shane Brennan, and Hai Tao.
\newblock Evaluating appearance models for recognition, reacquisition, and
  tracking.
\newblock In \emph{Proc. IEEE international workshop on performance evaluation
  for tracking and surveillance (PETS)}, volume~3, pages 1--7. Citeseer, 2007.

\bibitem[Gronskyte et~al.(2016)Gronskyte, Clemmensen, Hviid, and
  Kulahci]{gronskyte2016monitoring}
Ruta Gronskyte, Line~Harder Clemmensen, Marchen~Sonja Hviid, and Murat Kulahci.
\newblock Monitoring pig movement at the slaughterhouse using optical flow and
  modified angular histograms.
\newblock \emph{Biosystems Engineering}, 141:\penalty0 19--30, 2016.

\bibitem[Gupta et~al.(2021)Gupta, Gavves, and Smeulders]{gupta2021icpr}
Deepak~K. Gupta, Efstratios Gavves, and Arnold W.~M. Smeulders.
\newblock Tackling occlusion in siamese tracking with structured dropouts.
\newblock In \emph{2020 25th International Conference on Pattern Recognition
  (ICPR)}, pages 5804--5811, 2021.
\newblock \doi{10.1109/ICPR48806.2021.9412120}.

\bibitem[He et~al.(2016)He, Zhang, Ren, and Sun]{he2016deep}
Kaiming He, Xiangyu Zhang, Shaoqing Ren, and Jian Sun.
\newblock Deep residual learning for image recognition.
\newblock In \emph{Proceedings of the IEEE conference on computer vision and
  pattern recognition}, pages 770--778, 2016.

\bibitem[He et~al.(2017)He, Gkioxari, Doll{\'a}r, and Girshick]{he2017mask}
Kaiming He, Georgia Gkioxari, Piotr Doll{\'a}r, and Ross Girshick.
\newblock Mask r-cnn.
\newblock In \emph{Proceedings of the IEEE international conference on computer
  vision}, pages 2961--2969, 2017.

\bibitem[Kashiha et~al.(2013)Kashiha, Bahr, Ott, Moons, Niewold, {\"O}dberg,
  and Berckmans]{kashiha2013automatic}
Mohammadamin Kashiha, Claudia Bahr, Sanne Ott, Christel~PH Moons, Theo~A
  Niewold, Frank~O {\"O}dberg, and Daniel Berckmans.
\newblock Automatic identification of marked pigs in a pen using image pattern
  recognition.
\newblock \emph{Computers and electronics in agriculture}, 93:\penalty0
  111--120, 2013.

\bibitem[Kim et~al.(2010)Kim, Kim, and Park]{kim2010animal}
So-Hyeon Kim, Do-Hyeun Kim, and Hee-Dong Park.
\newblock Animal situation tracking service using rfid, gps, and sensors.
\newblock In \emph{2010 Second International Conference on Computer and Network
  Technology}, pages 153--156. IEEE, 2010.

\bibitem[Kuhn(1955)]{kuhn1955hungarian}
Harold~W Kuhn.
\newblock The hungarian method for the assignment problem.
\newblock \emph{Naval research logistics quarterly}, 2\penalty0 (1-2):\penalty0
  83--97, 1955.

\bibitem[Kuipers et~al.(2020)Kuipers, Arya, and Gupta]{kuipers2020eccvw}
Thijs~P. Kuipers, Devanshu Arya, and Deepak~K. Gupta.
\newblock Hard occlusions in visual object tracking.
\newblock In \emph{Computer Vision -- ECCV 2020 Workshops}, pages 299--314,
  2020.

\bibitem[Kulikov et~al.(2014)Kulikov, Khotskin, Nikitin, Lankin, Kulikov, and
  Trapezov]{kulikov2014application}
Victor~A Kulikov, Nikita~V Khotskin, Sergey~V Nikitin, Vasily~S Lankin,
  Alexander~V Kulikov, and Oleg~V Trapezov.
\newblock Application of {3-D} imaging sensor for tracking minipigs in the open
  field test.
\newblock \emph{Journal of neuroscience methods}, 235:\penalty0 219--225, 2014.

\bibitem[Lin et~al.(2017)Lin, Goyal, Girshick, He, and
  Doll{\'a}r]{lin2017focal}
Tsung-Yi Lin, Priya Goyal, Ross Girshick, Kaiming He, and Piotr Doll{\'a}r.
\newblock Focal loss for dense object detection.
\newblock In \emph{Proceedings of the IEEE international conference on computer
  vision}, pages 2980--2988, 2017.

\bibitem[Liu et~al.(2021)Liu, Lin, Cao, Hu, Wei, Zhang, Lin, and
  Guo]{liu2021swin}
Ze~Liu, Yutong Lin, Yue Cao, Han Hu, Yixuan Wei, Zheng Zhang, Stephen Lin, and
  Baining Guo.
\newblock Swin transformer: Hierarchical vision transformer using shifted
  windows.
\newblock \emph{arXiv preprint arXiv:2103.14030}, 2021.

\bibitem[Magee and Boyle(2000)]{magee2000}
Derek~R. Magee and Roger~D. Boyle.
\newblock Detecting lameness in livestock using ‘re-sampling condensation’
  and ‘multi-stream cyclic hidden markov models’.
\newblock In \emph{BMVC}, pages 564--586, 2000.

\bibitem[Mayer et~al.(2004)Mayer, Ellis, and Taylor]{mayer2004cattle}
Kevin Mayer, Keith Ellis, and Ken Taylor.
\newblock Cattle health monitoring using wireless sensor networks.
\newblock In \emph{Proceedings of the Communication and Computer Networks
  Conference (CCN 2004)}, pages 8--10. ACTA Press, 2004.

\bibitem[Meinhardt et~al.(2021)Meinhardt, Kirillov, Leal-Taixe, and
  Feichtenhofer]{meinhardt2021trackformer}
Tim Meinhardt, Alexander Kirillov, Laura Leal-Taixe, and Christoph
  Feichtenhofer.
\newblock Trackformer: Multi-object tracking with transformers.
\newblock \emph{arXiv preprint arXiv:2101.02702}, 2021.

\bibitem[Milletari et~al.(2016)Milletari, Navab, Ahmadi, and
  V-net]{milletari2016fully}
F~Milletari, N~Navab, SAV Ahmadi, and V-net.
\newblock Fully convolutional neural networks for volumetric medical image
  segmentation.
\newblock In \emph{Proceedings of the 2016 Fourth International Conference on
  3D Vision (3DV)}, pages 565--571, 2016.

\bibitem[Mittek et~al.(2016)Mittek, Psota, P{\'e}rez, Schmidt, and
  Mote]{mittek2016health}
Mateusz Mittek, Eric~T Psota, Lance~C P{\'e}rez, Ty~Schmidt, and Benny Mote.
\newblock Health monitoring of group-housed pigs using depth-enabled
  multi-object tracking.
\newblock In \emph{Proceedings of Int Conf Pattern Recognit, Workshop on Visual
  observation and analysis of Vertebrate And Insect Behavior}, 2016.

\bibitem[Nasirahmadi et~al.(2016)Nasirahmadi, Hensel, Edwards, and
  Sturm]{nasirahmadi2016automatic}
Abozar Nasirahmadi, Oliver Hensel, Sandra~A Edwards, and Barbara Sturm.
\newblock Automatic detection of mounting behaviours among pigs using image
  analysis.
\newblock \emph{Computers and Electronics in Agriculture}, 124:\penalty0
  295--302, 2016.

\bibitem[Nasirahmadi et~al.(2017)Nasirahmadi, Edwards, and
  Sturm]{nasirahmadi2017implementation}
Abozar Nasirahmadi, Sandra~A Edwards, and Barbara Sturm.
\newblock Implementation of machine vision for detecting behaviour of cattle
  and pigs.
\newblock \emph{Livestock Science}, 202:\penalty0 25--38, 2017.

\bibitem[Neethirajan(2017)]{neethirajan2017recent}
Suresh Neethirajan.
\newblock Recent advances in wearable sensors for animal health management.
\newblock \emph{Sensing and Bio-Sensing Research}, 12:\penalty0 15--29, 2017.

\bibitem[Nilsson et~al.(2015)Nilsson, Herlin, Ard{\"o}, Guzhva, {\AA}str{\"o}m,
  and Bergsten]{nilsson2015development}
Mikael Nilsson, AH~Herlin, H{\aa}kan Ard{\"o}, O~Guzhva, Karl {\AA}str{\"o}m,
  and C~Bergsten.
\newblock Development of automatic surveillance of animal behaviour and welfare
  using image analysis and machine learned segmentation technique.
\newblock \emph{Animal}, 9\penalty0 (11):\penalty0 1859--1865, 2015.

\bibitem[Oczak et~al.(2014)Oczak, Viazzi, Ismayilova, Sonoda, Roulston, Fels,
  Bahr, Hartung, Guarino, Berckmans, et~al.]{oczak2014classification}
Maciej Oczak, Stefano Viazzi, Gunel Ismayilova, Lilia~T Sonoda, Nancy Roulston,
  Michaela Fels, Claudia Bahr, J{\"o}rg Hartung, Marcella Guarino, Daniel
  Berckmans, et~al.
\newblock Classification of aggressive behaviour in pigs by activity index and
  multilayer feed forward neural network.
\newblock \emph{Biosystems Engineering}, 119:\penalty0 89--97, 2014.

\bibitem[O’shaughnessy et~al.(2012)O’shaughnessy, Peters, Donham, Taylor,
  Altmaier, and Kelly]{o2012assessment}
Patrick O’shaughnessy, Thomas Peters, Kelley Donham, Craig Taylor, Ralph
  Altmaier, and Kevin Kelly.
\newblock Assessment of swine worker exposures to dust and endotoxin during hog
  load-out and power washing.
\newblock \emph{Annals of occupational hygiene}, 56\penalty0 (7):\penalty0
  843--851, 2012.

\bibitem[Porto et~al.(2014)Porto, Arcidiacono, Giummarra, Anguzza, and
  Cascone]{porto2014localisation}
SMC Porto, C~Arcidiacono, A~Giummarra, U~Anguzza, and G~Cascone.
\newblock Localisation and identification performances of a real-time location
  system based on ultra wide band technology for monitoring and tracking dairy
  cow behaviour in a semi-open free-stall barn.
\newblock \emph{Computers and Electronics in Agriculture}, 108:\penalty0
  221--229, 2014.

\bibitem[Psota et~al.(2019)Psota, Mittek, P{\'e}rez, Schmidt, and
  Mote]{psota2019multi}
Eric~T Psota, Mateusz Mittek, Lance~C P{\'e}rez, Ty~Schmidt, and Benny Mote.
\newblock Multi-pig part detection and association with a fully-convolutional
  network.
\newblock \emph{Sensors}, 19\penalty0 (4):\penalty0 852, 2019.

\bibitem[Ruiz-Garcia et~al.(2009)Ruiz-Garcia, Lunadei, Barreiro, and
  Robla]{ruiz2009review}
Luis Ruiz-Garcia, Loredana Lunadei, Pilar Barreiro, and Ignacio Robla.
\newblock A review of wireless sensor technologies and applications in
  agriculture and food industry: state of the art and current trends.
\newblock \emph{sensors}, 9\penalty0 (6):\penalty0 4728--4750, 2009.

\bibitem[Schleppe et~al.(2010)Schleppe, Lachapelle, Booker, and
  Pittman]{schleppe2010challenges}
JB~Schleppe, G~Lachapelle, CW~Booker, and T~Pittman.
\newblock Challenges in the design of a {GNSS} ear tag for feedlot cattle.
\newblock \emph{Computers and Electronics in Agriculture}, 70\penalty0
  (1):\penalty0 84--95, 2010.

\bibitem[Schwager et~al.(2007)Schwager, Anderson, Butler, and
  Rus]{schwager2007robust}
Mac Schwager, Dean~M Anderson, Zack Butler, and Daniela Rus.
\newblock Robust classification of animal tracking data.
\newblock \emph{Computers and Electronics in Agriculture}, 56\penalty0
  (1):\penalty0 46--59, 2007.

\bibitem[Stavrakakis et~al.(2015)Stavrakakis, Li, Guy, Morgan, Ushaw, Johnson,
  and Edwards]{stavrakakis2015validity}
Sophia Stavrakakis, Wei Li, Jonathan~H Guy, Graham Morgan, Gary Ushaw, Garth~R
  Johnson, and Sandra~A Edwards.
\newblock Validity of the microsoft kinect sensor for assessment of normal
  walking patterns in pigs.
\newblock \emph{Computers and Electronics in Agriculture}, 117:\penalty0 1--7,
  2015.

\bibitem[Sun et~al.(2020)Sun, Jiang, Zhang, Xie, Cao, Hu, Kong, Yuan, Wang, and
  Luo]{sun2020transtrack}
Peize Sun, Yi~Jiang, Rufeng Zhang, Enze Xie, Jinkun Cao, Xinting Hu, Tao Kong,
  Zehuan Yuan, Changhu Wang, and Ping Luo.
\newblock Transtrack: Multiple-object tracking with transformer.
\newblock \emph{arXiv preprint arXiv:2012.15460}, 2020.

\bibitem[Taylor et~al.(1986)]{taylor1986pig}
David~J Taylor et~al.
\newblock \emph{Pig diseases.}
\newblock Number Edition 4. Dr. DJ Taylor, 31 North Birbiston Road, 1986.

\bibitem[Ter-Sarkisov et~al.(2018)Ter-Sarkisov, Ross, Kelleher, Earley, and
  Keane]{ter2018beef}
Aram Ter-Sarkisov, Robert Ross, John Kelleher, Bernadette Earley, and Michael
  Keane.
\newblock Beef cattle instance segmentation using fully convolutional neural
  network.
\newblock \emph{BMVC}, 2018.

\bibitem[Voigtlaender et~al.(2019)Voigtlaender, Krause, Osep, Luiten, Sekar,
  Geiger, and Leibe]{voigtlaender2019mots}
Paul Voigtlaender, Michael Krause, Aljosa Osep, Jonathon Luiten, Berin
  Balachandar~Gnana Sekar, Andreas Geiger, and Bastian Leibe.
\newblock Mots: Multi-object tracking and segmentation.
\newblock In \emph{Proceedings of the IEEE/CVF Conference on Computer Vision
  and Pattern Recognition}, pages 7942--7951, 2019.

\bibitem[Voulodimos et~al.(2010)Voulodimos, Patrikakis, Sideridis, Ntafis, and
  Xylouri]{voulodimos2010complete}
Athanasios~S Voulodimos, Charalampos~Z Patrikakis, Alexander~B Sideridis,
  Vasileios~A Ntafis, and Eftychia~M Xylouri.
\newblock A complete farm management system based on animal identification
  using rfid technology.
\newblock \emph{Computers and electronics in agriculture}, 70\penalty0
  (2):\penalty0 380--388, 2010.

\bibitem[Wallinga(2018)]{wallinga2018better}
David Wallinga.
\newblock Better bacon why it’s high time the {US} pork industry stopped
  pigging out on antibiotics, 2018.

\bibitem[Wang et~al.(2020)Wang, Mohamed, Le, Liu, Xiao, Mahadeokar, Huang,
  Tjandra, Zhang, Zhang, et~al.]{wang2020transformer}
Yongqiang Wang, Abdelrahman Mohamed, Due Le, Chunxi Liu, Alex Xiao, Jay
  Mahadeokar, Hongzhao Huang, Andros Tjandra, Xiaohui Zhang, Frank Zhang,
  et~al.
\newblock Transformer-based acoustic modeling for hybrid speech recognition.
\newblock In \emph{ICASSP 2020-2020 IEEE International Conference on Acoustics,
  Speech and Signal Processing (ICASSP)}, pages 6874--6878. IEEE, 2020.

\bibitem[Wedin et~al.(2018)Wedin, Baxter, Jack, Futro, and
  D’Eath]{wedin2018early}
Maya Wedin, Emma~M Baxter, Mhairi Jack, Agnieszka Futro, and Richard~B
  D’Eath.
\newblock Early indicators of tail biting outbreaks in pigs.
\newblock \emph{Applied animal behaviour science}, 208:\penalty0 7--13, 2018.

\bibitem[Wu et~al.(2019{\natexlab{a}})Wu, Zhao, Zhou, and Lu]{wu2019multi}
SiFeng Wu, XueBin Zhao, Hao Zhou, and Jun Lu.
\newblock Multi object tracking based on detection with deep learning and
  hierarchical clustering.
\newblock In \emph{2019 IEEE 4th International Conference on Image, Vision and
  Computing (ICIVC)}, pages 367--370. IEEE, 2019{\natexlab{a}}.

\bibitem[Wu et~al.(2019{\natexlab{b}})Wu, Kirillov, Massa, Lo, and
  Girshick]{wu2019detectron2}
Yuxin Wu, Alexander Kirillov, Francisco Massa, Wan-Yen Lo, and Ross Girshick.
\newblock Detectron2.
\newblock \url{https://github.com/facebookresearch/detectron2},
  2019{\natexlab{b}}.

\bibitem[Zhao et~al.(2021)Zhao, Okada, and Inaba]{zhao2021trtr}
Moju Zhao, Kei Okada, and Masayuki Inaba.
\newblock {TrTr}: Visual tracking with transformer.
\newblock \emph{arXiv preprint arXiv:2105.03817}, 2021.

\bibitem[Zhu et~al.(2020)Zhu, Su, Lu, Li, Wang, and Dai]{zhu2020deformable}
Xizhou Zhu, Weijie Su, Lewei Lu, Bin Li, Xiaogang Wang, and Jifeng Dai.
\newblock Deformable {DETR}: Deformable transformers for end-to-end object
  detection.
\newblock \emph{ICLR 2020}, 2020.

\end{thebibliography}
\end{document}